\pdfoutput=1

\documentclass[11pt]{article}

\usepackage[preprint]{acl}

\usepackage{times}
\usepackage{latexsym}

\usepackage[T1]{fontenc}

\usepackage[utf8]{inputenc}

\usepackage{microtype}

\usepackage{inconsolata}

\usepackage{graphicx}
\usepackage{tabularx}
\usepackage{array, multirow}
\usepackage{booktabs}
\usepackage{caption}
\usepackage{url}

\usepackage{amsmath,amsfonts,bm}

\def\eqref#1{equation~\ref{#1}}

\def\1{\bm{1}}

\DeclareMathAlphabet{\mathsfit}{\encodingdefault}{\sfdefault}{m}{sl}
\SetMathAlphabet{\mathsfit}{bold}{\encodingdefault}{\sfdefault}{bx}{n}

\title{A Dual-Axis Taxonomy of Knowledge Editing for LLMs: \\ From Mechanisms to Functions}

\author{
  Amir Mohammad Salehoof,
  Ali Ramezani,
  Yadollah Yaghoobzadeh,
  Majid Nili Ahmadabadi \\
  Department of Electrical and Computer Engineering \\
  University of Tehran \\
  \texttt{\{a.m.salehoof, ali.ramezani.96, y.yaghoobzadeh, mnili\}@ut.ac.ir}
}

\begin{document}
\maketitle
\begin{abstract}
Large language models (LLMs) acquire vast knowledge from large text corpora, but this information can become outdated or inaccurate. Since retraining is computationally expensive, knowledge editing offers an efficient alternative—modifying internal knowledge without full retraining. These methods aim to update facts precisely while preserving the model’s overall capabilities.

While existing surveys focus on the \textit{mechanism} of editing (e.g., parameter changes vs. external memory), they often overlook the \textit{function} of the knowledge being edited. This survey introduces a novel, complementary \textbf{function-based taxonomy} to provide a more holistic view. We examine how different mechanisms apply to various knowledge types—\textbf{factual, temporal, conceptual, commonsense, and social}—highlighting how editing effectiveness depends on the nature of the target knowledge.

By organizing our review along these two axes, we map the current landscape, outline the strengths and limitations of existing methods, define the problem formally, survey evaluation tasks and datasets, and conclude with open challenges and future directions.

\end{abstract}

\section{Introduction}
\label{sec:introduction} 

Large language models (LLMs) have shown remarkable abilities in understanding and generating human-like text \citep{gpt3, gpt4, palm2, llama, zhao2023survey}. However, keeping them relevant and correcting errors efficiently remains a challenge. Retraining entire models is computationally expensive, prompting interest in model editing \citep{ENN, factual}, which enables targeted updates while preserving overall functionality.

\begin{figure}[t] 
    \centering
    \includegraphics[width=\linewidth]{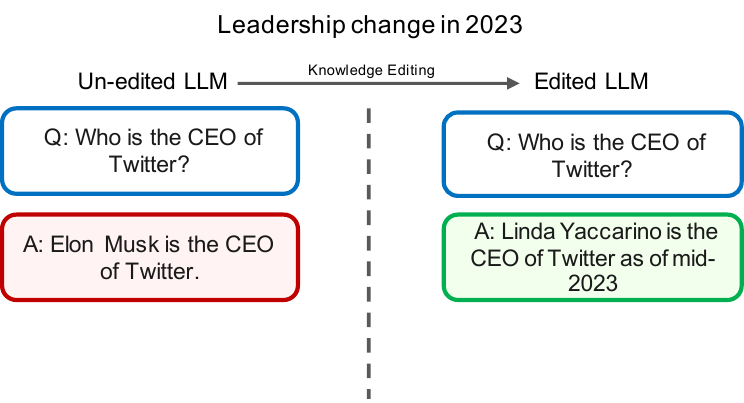}
    \caption{An example of Knowledge Editing illustrating efficient and localized knowledge updates in LLMs.}
    \label{fig:ke}
\end{figure}

As shown in Figure \ref{fig:ke}, knowledge editing aims to correct specific information in a model. When an LLM gives an incorrect output, an editor adjusts the model to produce a factual response, with changes localized to avoid affecting unrelated knowledge.

Although various KE methods have emerged \citep{factual, MEMIT, ROME, ENN, patcher}, most surveys classify them by their mechanisms—modifying parameters or adding external modules. This overlooks an essential aspect: the type of knowledge being edited. Techniques that work for simple facts (e.g., capital cities) may fall short with complex knowledge like commonsense reasoning or social biases.

This survey addresses the gap by proposing a novel, complementary function-based taxonomy. We argue that understanding KE requires examining the type of knowledge being edited. By classifying methods by the functional knowledge they target—factual, temporal, conceptual, commonsense, and social—we reveal the unique challenges and limitations of current approaches. This framework offers a more holistic basis for evaluating and advancing the field.

To guide the reader, Section \ref{sec:problem} defines the problem and outlines key properties of an ideal editor. Section \ref{sec:methods} introduces our dual-axis taxonomy, covering both mechanism- and function-based perspectives. Section \ref{sec:datasets} surveys evaluation tasks and datasets, and Section \ref{sec:challenges} highlights open challenges and future directions.

\section{Knowledge Editing}
\label{sec:problem}

\textbf{Knowledge editing} (KE), also known as \textbf{model editing}, was first introduced by \citet{ENN}. The core objective is to correct a model’s error on a specific instance while preserving its overall behavior. For a base model $f_\theta$ and a specific edit request—an input-output pair $(x_e, y_e)$ where the model's current output is incorrect ($f_\theta(x_e) \neq y_e$)—the goal is to produce an edited model, $f_{\theta_e}$, that satisfies the request ($f_{\theta_e}(x_e) = y_e$) \citep{MEND, yao2023editing}.

The central challenge of KE lies in achieving this correction with precision. An ideal editor must make changes that are both specific enough to avoid unintended side effects and general enough to be robust. To formalize this, we define two disjoint sets of inputs:
\begin{itemize}
    \item \textbf{Edit Scope} $I(x_e, y_e)$: The set of all inputs to which the new fact should apply. This includes the original input $x_e$ and all its semantic paraphrases (e.g., different ways of asking the same question).
    \item \textbf{Out-of-Scope} $O(x_e, y_e)$: The set of all other inputs, which should remain completely unaffected by the edit.
\end{itemize}

A successful KE method, therefore, must satisfy the condition outlined in Equation \ref{eq:editing}, which states that the edited model should produce the new target output for all in-scope inputs and revert to its original behavior for all out-of-scope inputs \citep{yao2023editing}.

\begin{equation}
f_{\theta_e}(x) =
\begin{cases}
y_e & \text{if } x \in I(x_e, y_e), \\
f_\theta(x) & \text{if } x \in O(x_e, y_e)
\end{cases}
\label{eq:editing}
\end{equation}

\subsection{General Metrics}
To measure how well a given method approximates this ideal, the literature has established four general properties: Reliability, Generality, Locality, and Efficiency. While these metrics provide a foundational assessment, we will later introduce more specialized, function-specific metrics in our analysis of different knowledge types

\begin{itemize}
    \item \textbf{Reliability:}
    Reliability measures if the edit was successful for the specific input it was given. It is the most fundamental property of a successful edit \citep{patcher,factual,ROME}.
    Formally, it is the success rate on the original edit pair $(x_e, y_e)$.
    \begin{equation}
    \mathbb{E}_{x_{\mathrm{e}}^{\prime},y_{\mathrm{e}}^{\prime} \sim\{\left(x_{\mathrm{e}}, y_{\mathrm{e}}\right)\}}
    \mathbb{I}\bigl\{\operatorname{argmax}_y f_{\theta_e}\left(y \mid x_{\mathrm{e}}^{\prime}\right)=y_{\mathrm{e}}^{\prime}\bigr\}
    \label{eq:reliability}
    \end{equation}
    In simple terms, this metric asks: After the edit, does the model now provide the correct target answer for the original prompt?

    \item \textbf{Generality:}
    Generality (or Generalization) measures whether the edit propagates to other semantically equivalent inputs that fall within the edit scope $I(x_e, y_e)$. This is typically evaluated on a set of paraphrases or ``neighboring instances,'' denoted as $N(x_e, y_e)$, to ensure the updated knowledge is robust and not just a superficial fix.
    \begin{equation}
    \mathbb{E}_{x_{\mathrm{e}}^{\prime}, y_{\mathrm{e}}^{\prime} \sim N\left(x_{\mathrm{e}}, y_{\mathrm{e}}\right)}
    \mathbb{I}\bigl\{\operatorname{argmax}_y f_{\theta_e}\left(y \mid x_{\mathrm{e}}^{\prime}\right)=y_{\mathrm{e}}^{\prime}\bigr\}
    \label{eq:generality}
    \end{equation}
    This metric essentially asks: Does the edit also apply to different phrasings of the same question?

    \item \textbf{Locality:}
    Locality, also known as \textbf{specificity} \citep{yao2023editing}, measures whether the edit has had unintended effects on unrelated knowledge (i.e., on inputs in the out-of-scope set $O(x_e, y_e)$). High locality is critical for preserving the model's overall integrity.
    \begin{equation}
    \mathbb{E}_{x_{\mathrm{e}}^{\prime}, y_{\mathrm{e}}^{\prime} \sim O\left(x_{\mathrm{e}}, y_{\mathrm{e}}\right)}
    \mathbb{I}\Bigl\{f_{\theta_e}\left(y \mid x_{\mathrm{e}}^{\prime}\right) = f_\theta\left(y \mid x_{\mathrm{e}}^{\prime}\right)\Bigr\}
    \label{eq:locality}
    \end{equation}
    In other words, this checks that for unrelated inputs, the edited model’s output distribution is identical to the original model’s, ensuring there are no negative side effects.

    \item \textbf{Efficiency:}
    The KE method must be efficient in terms of computational resources, including both time and memory consumption \citep{mazzia2024survey}. Efficiency is especially crucial for practical applications involving large-scale models or streams of sequential edits.
\end{itemize}


\section{Dual-Axis Taxonomy:}
\label{sec:methods}

To provide a comprehensive overview of Knowledge Editing (KE), we analyze current techniques along two orthogonal axes: the \textit{mechanism} used to alter the model and the \textit{function} of the knowledge being targeted. This dual-perspective approach is essential because a method's effectiveness is defined by both its technical implementation and the nature of the problem it is intended to solve.

We begin in Section~\ref{sec:mechanism_based} by reviewing the primary editing mechanisms, which are broadly categorized as either modifying the model's parameters or preserving them. Then, in Section~\ref{sec:function_based}, we introduce our novel function-based taxonomy to analyze how these mechanisms are applied to increasingly complex types of knowledge.

\subsection{Mechanism-Based Editing: How Is the Model Altered?}
\label{sec:mechanism_based}

KE techniques are most commonly distinguished by \textbf{how} they alter a model's behavior. The central choice is whether to directly change the LLM's internal weights or to augment the model with an external component that intercepts or guides its outputs at inference time. The field's pioneering studies, such as ROME~\citep{ROME}, MEMIT~\citep{MEMIT}, and MEND~\citep{MEND}, were developed to correct discrete factual inaccuracies like \texttt{(Paris, capital\_of, France)} and thus established many of the foundational mechanisms discussed here.

\subsubsection{Parameter-Modifying Methods}
\label{sssec:param_mod}

These methods directly modify a model’s internal weights to encode new or corrected knowledge and fall into two main categories.

\textbf{Locate-then-Edit methods} aim for surgical precision by identifying and updating specific neurons or layers responsible for a piece of knowledge. \textbf{ROME}~\citep{ROME} uses causal mediation analysis to locate factual associations in transformer feed-forward layers, applying constrained optimization for edits. \textbf{MEMIT}~\citep{MEMIT} scales this by editing thousands of facts via efficient rank-one updates to the same layer type. \textbf{PMET}~\citep{PMET} extends this approach by including attention layers for finer control. While precise, these methods are less tested on non-factual knowledge, and the reliability of causal localization remains uncertain~\citep{surprising}.

\textbf{Hypernetwork/Meta-Learning approaches} use a separate model to predict weight updates. \textbf{MEND}~\citep{MEND} trains a hypernetwork that converts gradients into low-rank updates. \textbf{MALMEN}~\citep{MALMEN} improves scalability by framing update prediction as a least-squares problem. Though flexible, these methods can be sensitive to domain shifts and cumulative edits.

\subsubsection{Parameter-Preserving Methods}
\label{sssec:param_preserve}

These methods keep the base LLM's weights frozen and instead modify its output behavior at inference time, prioritizing stability and reversibility.

\textbf{Memory-Based approaches} store new facts in external memory. \textbf{SERAC}~\citep{SERAC} uses a classifier to decide whether to rely on the base model or retrieve a counterfactual edit. \textbf{IKE}~\citep{IKE} and \textbf{MeLLo}~\citep{MeLLo} retrieve relevant examples to serve as in-context demonstrations, effectively editing model behavior without weight changes.

\textbf{Neuron-Augmented methods} insert trainable components into the architecture. \textbf{T-Patcher}~\citep{patcher} assigns a dedicated "patch" neuron per edit, activated as needed. \textbf{GRACE}~\citep{GRACE} caches corrective activations in a codebook to support sequential edits. \textbf{CaliNet}~\citep{calinet} adds small, tunable modules for factual calibration. These methods trade deep integration for locality, offering strong stability, reversibility, and minimal side effects—key benefits for real-world deployment.

A detailed performance comparison of these mechanism-based methods on foundational benchmarks is provided in Appendix~\ref{app:results}.

\subsection{Function-Based Editing: What Kind of Knowledge Is Targeted?}
\label{sec:function_based}

While understanding the \textit{how} of editing is crucial, a full picture only emerges when we also consider \textit{what} is being edited. Early work focused almost exclusively on static, factual triples. However, as the field has matured, researchers have begun tackling more complex knowledge types that present unique challenges. In this section, we analyze recent work through this functional lens, systematically connecting each problem back to the mechanisms introduced previously.

\subsubsection{Temporal Knowledge}
\label{sssec:temporal}

We begin our exploration of knowledge types with \textbf{temporal knowledge}, a natural extension of static factual editing. Real-world knowledge often evolves (e.g., “The president of the USA is Joe Biden”), presenting a challenge for static models that require updates that reflect new information without erasing historical context. Editing this knowledge type introduces a unique challenge centered on \textbf{Locality}, as the primary goal is to update facts without corrupting the model's memory of relevant historical information. To address this, \textbf{METO}~\citep{Yin2023HistoryMT} introduces the Temporal Knowledge Editing (TKE) task and a corresponding benchmark, \textbf{ATOKE}. This method enhances existing \textbf{locate-then-edit} approaches like ROME and MEMIT with a multi-editing mechanism and time-sensitive objective, enabling joint optimization over both current and historical knowledge. To evaluate performance, the authors introduce several specialized metrics that map to the general principles of KE. Edit success is measured with a \textbf{Current Question Score (CES/CRS)}, which functions as a direct test of \textbf{Reliability}, and a \textbf{Paraphrase Score (CES-P)} to ensure \textbf{Generality}. Most critically, they use a \textbf{Historical Question Score (HES/HRS)} to assess if the model preserves the original fact as historical context. This offers a more nuanced measure of \textbf{Locality}, focusing on the preservation of relevant temporal facts rather than just the absence of unrelated errors~\citep{Yin2023HistoryMT}. Despite improvements, reasoning over relative temporal expressions and maintaining coherence across long factual chains remain open challenges.

\subsubsection{Conceptual Knowledge}
\label{sssec:conceptual}

\textbf{Conceptual knowledge} includes abstract definitions and category-level relationships, such as the definition of “mammal” or the criteria for being a “bachelor.” For this knowledge type, the central challenge is achieving a deep, structural form of \textbf{Generality}, where an edit to an abstract definition must consistently propagate 'top-down' to all of its member instances while maintaining semantic coherence.

\begin{itemize}
    \item \textbf{ConceptEdit}~\citep{Wang2024EditingCK} pioneers this task by establishing the first benchmark to evaluate how existing methods handle conceptual edits. Instead of proposing a new technique, it assesses standard \textbf{Parameter-Modifying} approaches, revealing a critical gap: while methods like ROME and MEMIT achieve high \textbf{Reliability} in changing a concept's definition, they demonstrate poor structural \textbf{Generality} in propagating these changes to instance-level knowledge. The paper introduces two tailored metrics to capture this: \textbf{Concept Consistency} as a nuanced measure of Reliability, and \textbf{Instance Change} to directly evaluate this top-down Generality.

    \item \textbf{RelEdit}~\citep{Niu2025RelEditEC} builds on prior work by arguing that evaluating conceptual edits requires moving beyond simple definition changes to assess the edit's impact on the model's \textbf{relational reasoning}. It introduces a more comprehensive benchmark, \textbf{RelEdit}, with a suite of new metrics designed to test these "ripple effects" on the relationships between both concepts and instances. These metrics provide a more fine-grained assessment of general KE principles: structural \textbf{Generality} is measured through metrics like \textit{Portability} (assessing if new instances correctly associate with the edited concept) and \textit{Alignment Belong/Compare} (checking for correct propagation through the conceptual hierarchy). \textbf{Locality} is specifically tested with \textit{Instance Locality}, which ensures unrelated instance-concept pairs remain unaffected. To address the propagation challenge identified by prior work, the paper proposes a non-parametric baseline, \textbf{MICE (Memory-based In-Context Editing)}, which uses an external memory and in-context learning. The finding that MICE outperforms traditional parameter-modifying methods on these complex reasoning tasks suggests that memory-based approaches are a promising direction for this field.
\end{itemize}

\subsubsection{Commonsense Knowledge}
\label{sssec:commonsense}

\textbf{Commonsense knowledge} encompasses intuitive, everyday reasoning about the physical and causal world (e.g., “Rain makes the ground wet”). Editing this knowledge type pushes the boundaries of both \textbf{Generality} and \textbf{Locality}. The challenge lies in propagating an edit through a web of informal, interconnected facts while preserving related but distinct concepts, requiring a more sophisticated evaluation framework. Unlike structured factual knowledge, it is often expressed in free-text and is distributed across a model's architecture, making it difficult to localize and edit. Early methods designed for single-token, triple-based facts thus face fundamental limitations in this domain.

To improve applicability in the commonsense domain, recent work has focused on adapting the \textbf{Locate-then-Edit} mechanism to handle this distributed knowledge:
\begin{itemize}
    \item \textbf{MEMITCSK}~\citep{Gupta2023EditingCS} extends its predecessor, MEMIT, to handle the unique challenges of \textit{commonsense knowledge}, which, unlike encyclopedic facts, often involves uncertainty and multiple plausible answers. The paper argues that for commonsense, plausibility judgments depend on the entire subject-verb-object triple. Accordingly, it improves MEMIT's \textbf{locate-then-edit} mechanism in two ways: (1) performing causal tracing and editing on subject, verb, and object tokens, and (2) using a more robust layer selection strategy based on the moving average of the Average Indirect Effect (AIE). To provide a more comprehensive evaluation, the paper introduces the \textbf{PROBE SET} benchmark, which includes specialized tests that map to general KE principles. \textbf{Locality} is measured via an \textit{Unaffected Neighborhood} (related but distinct facts that should not change). \textbf{Generality} is assessed through an \textit{Affected Neighborhood} (synonyms), \textit{Affected Paraphrases}, and, most notably, an \textit{Affected Reasoning} set, which tests if the edit propagates through a simple logical chain.
    
    \item \textbf{DEM}~\citep{Huang2024CommonsenseKE} addresses the challenge of editing free-text \textit{commonsense knowledge}, which differs from factual knowledge due to its multi-token nature and distributed storage. The authors first use a novel analysis method, \textbf{KLFT}, to demonstrate that commonsense knowledge is dispersed across both \textbf{MLP} and \textbf{Attention layers}, unlike factual knowledge which is more localized. Motivated by this finding, they propose a dynamics-aware editing mechanism. This method consists of two parts: (1) a \textbf{Dynamics-aware Module} that dynamically identifies the most relevant layers for each specific edit, rather than using a fixed location, and (2) a \textbf{Knowledge Editing Module} that jointly updates parameters in both the identified MLP and Attention layers. To support evaluation, the paper introduces the \textbf{CKEBench} dataset. It assesses performance using adapted metrics for free-text, including \textbf{Score} (for \textbf{Reliability}), \textbf{Specificity} (for \textbf{Locality}), and \textbf{Generalization}, all evaluated via GPT-4 similarity. It also introduces a new domain-specific \textbf{Commonsense} metric to verify the edit's underlying success.
\end{itemize}
Together, these methods illustrate diverse strategies for commonsense knowledge editing: from refined token-layer targeting (MEMITCSK), to dynamic structural localization (DEM). Nonetheless, editing commonsense remains an open challenge due to its contextuality, ambiguity, and distributed nature. Open questions include scaling to multilingual and multimodal settings, resolving conflicting edits, and preserving coherence across related concepts.

\subsubsection{Social Knowledge}
\label{sssec:social}

\textbf{Social knowledge} editing targets biased or harmful associations embedded in language models, such as gender stereotypes or toxic completions. In this domain, the critical challenge is balancing \textbf{Reliability} with \textbf{Locality}. The goal is to precisely remove a harmful association (\textbf{Reliability}) while rigorously preserving the model's useful knowledge and general capabilities (\textbf{Locality}), avoiding the common failure mode of corrupting valid information in the pursuit of fairness. While early debiasing approaches often relied on methods like prompt engineering, they typically lacked persistence and control. More recently, researchers have explored knowledge editing as an alternative paradigm—shifting the focus from output steering to direct modification of the model’s internal representations and parameters. This approach, distinct from alignment strategies like RLHF and DPO, enables more targeted and interpretable edits to the underlying knowledge responsible for social bias.

The following works illustrate three complementary strategies within this emerging paradigm, each adapting a different core mechanism:
\begin{itemize}
    \item \textbf{BIASEDIT}~\citep{Xu2025BiasEditDS} adapts the \textbf{Hypernetwork/Meta-Learning} approach for bias mitigation. Building on the MEND~\citep{MEND} architecture, it introduces editor hypernetworks trained to modify stereotype-related parameters. BIASEDIT proposes a pair of objectives that directly map to core KE principles. To ensure \textbf{Reliability}, it uses a \textit{debiasing loss} to equalize the likelihoods of stereotypical and anti-stereotypical contexts, with success measured by the \textbf{Stereotype Score (SS)}, which aims for an ideal value of 50\%. To maintain \textbf{Locality}, it employs a \textit{retention loss} to preserve the model's behavior on unrelated inputs (specifically, meaningless sentences). This is evaluated using the \textbf{Language Modeling Score (LMS)}, where a minimal change indicates that the model's general capabilities are unharmed. The paper also explicitly tests for \textbf{Generality} by evaluating the model on a synonym-augmented test set.
    \item \textbf{FAST}~\citep{Chen2024IdentifyingAM} addresses a key failure in existing debiasing work: that enforcing group-level parity often corrupts valid commonsense knowledge (e.g., making "mom" and "dad" biologically equivalent). It proposes a fine-grained approach analogous to \textbf{Neuron-Augmentation}. The framework first uses a contrastive method to \textit{localize} the single model layer most responsible for a specific bias. Then, it inserts a lightweight, trainable module called a \textbf{Fairness-Stamp (FAST)} at that location to perform a modular correction, while freezing the original model parameters. To evaluate this approach, the paper introduces the \textbf{BiaScope} benchmark with two new metrics. The \textbf{Retention Score (RS)} serves as a direct measure of \textbf{Locality}, quantifying how well the model preserves non-biased commonsense facts that should be unaffected. The \textbf{Paraphrase Stereotype Score (PS)} measures \textbf{Generality}, assessing if the debiasing effect extends to semantically similar, paraphrased sentences.
    \item \textbf{DINM}~\citep{Wang2024DetoxifyingLL} uses the \textbf{Locate-then-Edit} pipeline for detoxifying generative models from harmful behaviors triggered by adversarial prompts. Instead of tracing specific subject tokens, which is difficult in complex queries, DINM introduces a novel localization method. It identifies the "toxic layer" by finding the layer with the \textit{maximal hidden state difference} between a generated safe and unsafe response to the same query. It then fine-tunes only the parameters of the MLP components within this single toxic layer using a safety-aware objective. To evaluate this approach, the paper constructs the \textbf{SafeEdit} benchmark, which includes metrics that map directly to KE principles. \textbf{Reliability} is measured by \textbf{Defense Success (DS)} on the original adversarial prompt. \textbf{Generality} is assessed with a suite of \textbf{Defense Generalization (DG)} metrics that test the model against out-of-domain questions and attack prompts. Finally, \textbf{Locality} is evaluated by measuring the impact on general capabilities like \textbf{Fluency} and performance on downstream tasks such as \textbf{Knowledge QA} and \textbf{Summarization}.
\end{itemize}
Taken together, these works exemplify three complementary angles on editing social knowledge: parameter-space rewiring (BIASEDIT), activation-space probing and correction (FAST), and behavior-level detoxification through adversarial supervision (DINM). Each builds on a different base: BIASEDIT on MEND-style hypernetworks, FAST on contrastive localization and modular correction, and DINM on ROME-like causal tracing and editing. Yet, they also highlight shared challenges—maintaining general language ability, minimizing unintended interference, and adapting to multilingual or evolving social norms. These remain important directions for future research.

\section{Tasks and Datasets}
\label{sec:datasets}

Evaluating KE methods requires well-defined tasks and robust datasets that can assess the effectiveness of different editing techniques. Various tasks have been proposed to test how well models incorporate, retain, and generalize knowledge edits, with a strong emphasis on factual accuracy, consistency, and minimal unintended changes to unrelated knowledge \citep{KE_survey}.

\subsection{Tasks}
\label{ssec:tasks}

KE tasks evaluate how well a model integrates factual modifications while preserving existing knowledge. These tasks serve as benchmarks for measuring the effectiveness of different KE approaches. The primary tasks considered in KE research include:
\begin{itemize}
    \item \textbf{Fact-Checking (FC)}: Assessing the model’s ability to verify and correct factual claims based on external evidence or world knowledge. This includes static facts, time-sensitive claims, and social assertions (e.g., stereotypical or biased statements).
    \item \textbf{Question Answering (QA)}: Evaluating how well a model retrieves and updates factual, temporal, or commonsense knowledge in response to questions. This includes closed-book QA where models must reflect the most recent or correct version of edited knowledge.
    \item \textbf{Natural Language Generation (NLG)}: Testing whether edits are reflected in free-form outputs, including summaries, descriptions, or generative completions that involve time-sensitive, social, or conceptual facts.
\end{itemize}


\begin{table*}[t]
\centering
\caption{Summary of papers by knowledge type and their primary mechanism or contribution.}
\label{tab:functional_mechanism_table}
\renewcommand{\arraystretch}{1.2}
\begin{tabular}{lll}
\toprule
\textbf{Functional Knowledge} & \textbf{Paper(s)} & \textbf{Primary Mechanism / Contribution} \\
\midrule
\multirow{4}{*}{Factual} 
& ROME, MEMIT, PMET & Locate-then-Edit \\
& MeLLo, SERAC, IKE & Memory \\
& CaliNet, T-Patcher, GRACE & Neuron-Augmented \\
& MEND, MALMEN & Meta-Learning \\
\midrule
Temporal & METO & Locate-then-Edit \\
\midrule
\multirow{2}{*}{Conceptual} 
& ConceptEdit$^{*}$ & --- \\
& RelEdit & Memory-based / In-Context (MICE) \\
\midrule
\multirow{2}{*}{Commonsense} 
& MEMITCSK & Locate-then-Edit (Extension) \\
& DEM & Locate-then-Edit (Distributed) \\
\midrule
\multirow{3}{*}{Social} 
& BIASEDIT & Hypernetwork / Meta-Learning \\
& FAST & Neuron-Augmented \\
& DINM & Locate-then-Edit \\
\bottomrule
\end{tabular}
\par\noindent
\footnotesize{$^{*}$Note: `ConceptEdit` does not propose a new editing method but evaluates existing ones on its conceptual knowledge benchmark.}
\end{table*}

\subsection{Datasets}
\label{ssec:datasets_list}

A broad suite of public datasets evaluates KE across functional dimensions, from factual updates to bias mitigation. Table~\ref{tab:datasets-small} summarizes these benchmarks; full descriptions appear in Appendix~\ref{app:datasets}.

\paragraph{Factual and Temporal Knowledge.}
Factual editing is assessed using generation-based datasets like \textbf{zsRE} and \textbf{CounterFact}, which test precision on isolated updates. \textbf{ATOKE} and \textbf{MQuAKE} extend this by evaluating temporal consistency and multi-hop reasoning for evolving or interdependent facts.

\paragraph{Conceptual and Commonsense Knowledge.}
Editing abstract knowledge requires higher-order reasoning benchmarks. \textbf{ConceptEdit} targets structural changes in definitions and their downstream effects~\citep{Wang2024EditingCK}, while \textbf{RelEdit} evaluates edits' impact on \textit{relational reasoning}~\citep{Niu2025RelEditEC}. \textbf{CKEBench} and \textbf{AbstractATOMIC} assess generalization and plausibility in commonsense contexts~\citep{Huang2024CommonsenseKE}.

\paragraph{Social Bias and Safety.}
Socially aware editing is evaluated with benchmarks like \textbf{Wikibias} and \textbf{BiaScope}, which address stereotype correction. \textbf{SafeEdit} measures the ability to neutralize harmful outputs while preserving fluency.

\section{Challenges andFuture directions}
\label{sec:challenges}

Knowledge Editing (KE) has emerged as a crucial research area for refining and updating factual knowledge in LLMs. While significant progress has been made, several challenges remain unaddressed, and future research directions must focus on improving efficiency, scalability, and robustness. This section outlines key challenges and promising future directions in KE.

\subsection{Challenges}

\subsubsection{Balancing Locality and Generalization}
A central challenge in KE is balancing \textit{locality} (avoiding side effects) with \textit{generalization} (ensuring consistency across contexts), depending on the knowledge \textbf{function}. \textbf{Factual} edits require high locality to prevent corruption, while \textbf{conceptual} or \textbf{social} edits demand broader generalization. Future work must develop methods that \textit{adaptively balance} this tradeoff by knowledge type.

\subsubsection{The Need for Theoretical Foundations}
Most KE methods are empirical and lack predictability due to the absence of a formal framework for how LLMs store, retrieve, and modify knowledge. Advancing the field requires theoretical foundations rooted in \textit{information theory, interpretability, and optimization} to guide principled editing strategies.

\subsubsection{Scalability to Mass-Edits}
Scaling KE to thousands of edits introduces conflicts, especially across heterogeneous knowledge types (e.g., \textbf{commonsense} and \textbf{factual}). Addressing this demands \textit{scalable architectures, memory-efficient representations}, and \textit{multi-edit synchronization} to maintain consistency and efficiency.

\subsubsection{Moving Beyond Structured Knowledge}
Current KE methods focus on structured, triple-based facts, leaving a gap in editing \textit{unstructured} sources like news. This is especially limiting for \textbf{commonsense} and \textbf{social} knowledge. Future work should build end-to-end pipelines to \textit{extract, validate, and integrate} edits from raw text, along with more flexible evaluation benchmarks.

\begin{table}[t]
\centering
\small
\caption{Summary of KE datasets by knowledge type.}
\label{tab:datasets-small}
\renewcommand{\arraystretch}{1.1}
\begin{tabular}{@{}l l@{}}
\toprule
\textbf{Dataset} & \textbf{Type} \\
\midrule
\multicolumn{2}{@{}l}{\itshape\textbf{Generation-Based Datasets}} \\
zsRE & Factual \\
CounterFact & Factual \\
MQuAKE & Factual / Temporal \\
WikiGen & Factual \\
ATOKE & Temporal \\
CKEBench & Commonsense \\
AbsATOMIC & Conceptual / Commonsense \\
SafeEdit & Social \\
\midrule
\multicolumn{2}{@{}l}{\itshape\textbf{Classification-Based Datasets}} \\
FEVER & Factual \\
VitaminC & Factual \\
ConceptEdit & Conceptual \\
RelEdit & Conceptual \\
PROBE SET & Commonsense \\
Wikibias & Social \\
BiaScope & Social / Commonsense \\
SCOTUS &  Temporal \\
\bottomrule
\end{tabular}
\vspace{-1em}  
\end{table}

\subsection{Future Directions}

\subsubsection{Towards Optimization-Free and Runtime Editing}
Optimization-based KE is often too slow for real-time use. Future work should explore \textit{optimization-free methods}, such as in-context learning or memory-augmented models, enabling \textit{runtime knowledge adaptation} through dynamic user feedback without retraining.

\subsubsection{Automating the Discovery of Knowledge to Edit}
Current KE systems rely on manual error identification. Future approaches should \textit{automate edit discovery} from real-time knowledge streams using techniques like anomaly detection—essential for domains like \textit{healthcare} and \textit{finance}.

\subsubsection{Enhancing Robustness and Security}
KE introduces risks of malicious edits (e.g., \textit{biases, misinformation, backdoors}). Future work must develop \textit{verification, auditing}, and \textit{certification protocols} to ensure the security and trustworthiness of edited models.

\subsubsection{Developing Ethical and Fair Editing Frameworks}
Informed by \textbf{social knowledge editing} (see \ref{sssec:social}), fair KE must account for the ethical implications of deciding what to edit. Future work should build frameworks for \textit{transparency, community oversight}, and balancing factual accuracy with societal fairness.

\subsubsection{Creating Unified Evaluation Frameworks}
KE evaluation is currently fragmented across isolated benchmarks. A key direction is building \textbf{unified evaluation suites} that assess editors across diverse knowledge types, revealing tradeoffs (e.g., strong factual locality vs. weak conceptual generalization).

\section{Conclusion}
Maintaining the factual accuracy of LLMs as real-world information evolves is a persistent challenge. Knowledge Editing (KE) has emerged as an efficient solution, enabling targeted updates to an LLM’s internal knowledge without requiring costly full retraining.

This survey provided a comprehensive review of KE by analyzing the field along two orthogonal axes: the editing \textit{mechanism} and the knowledge \textit{function}. We categorized mechanisms into parameter-modifying and parameter-preserving approaches, then introduced our novel function-based taxonomy. This provides a holistic perspective by examining how these mechanisms apply to diverse knowledge types—from factual and temporal to conceptual, commonsense, and social—supplemented by an overview of the field's key properties, evaluation tasks, and datasets.

Despite remarkable progress, KE remains an evolving field. As we highlighted, future advancements must focus on developing adaptive, scalable, and secure editors. As LLMs become increasingly integrated into real-world applications, KE will be crucial for maintaining their reliability and adaptability, contributing to more dynamic, accurate, and ethically responsible AI systems.

\section*{Limitations}

The field of knowledge editing is evolving at an exceptional pace. While we have strived to provide a comprehensive overview, this survey represents a snapshot of research primarily published by mid-2025. New methods and preprints emerging during the review period may not be included.

Our primary contribution is a high-level taxonomic framework. To maintain this broad perspective, we prioritize the categorization and synthesis of different approaches over a deep, technical analysis of the implementation details of every individual method cited. Furthermore, our scope is strictly focused on knowledge editing, and we do not provide a detailed comparison with related but distinct fields such as continual learning or parameter-efficient fine-tuning.

Finally, this survey is a work of analysis and does not introduce new empirical results. All performance metrics discussed or presented (e.g., in Appendix A) are reported from the original publications. We did not re-run experiments to perform a controlled, head-to-head comparison of methods under a single, unified environment, as this is beyond the scope of a survey.


\section*{Acknowledgments}

We used OpenAI’s ChatGPT-4o to support this work. Specifically, we used it for grammar correction, clarity improvement, and literature search suggestions. All technical contributions, ideas, and conclusions remain entirely our own.


\bibliography{references}
\clearpage
\appendix

\section{Performance of Mechanism-Based Editors}\label{app:results}

To evaluate the performance of existing knowledge editing (KE) techniques, we summarize reported results on two benchmark datasets: \textbf{ZsRE} and \textbf{CounterFact}. These evaluations focus on three core metrics—\textit{reliability}, \textit{generalization}, and \textit{locality}—across different model architectures, specifically T5-XL and GPT-J. Note that we only include mechanism-based methods in Table~\ref{tab:ke-results}, as function-based approaches are evaluated on diverse and non-overlapping datasets, preventing fair comparison.

As shown in Table~\ref{tab:ke-results}, different methods demonstrate varying strengths. On ZsRE with T5-XL, SERAC achieves the highest reliability and generalization, while MEND provides the strongest locality. On GPT-J, IKE excels in reliability and generalization, whereas MEMIT achieves the best locality.

For the CounterFact dataset, SERAC again performs best in reliability and generalization for T5-XL, while KE surprisingly achieves the top score in locality. With GPT-J, T-Patcher stands out with perfect reliability, while SERAC leads in generalization and locality.

These results highlight that no single method dominates across all criteria. Techniques like SERAC and MEMIT provide robust general-purpose editing, while others such as IKE and KE offer targeted strengths depending on the task and architecture \citep{yao2023editing}.

\begin{table*}[ht]
\centering
\caption{Performance comparison of knowledge editing methods across datasets (ZsRE and CounterFact) and models (T5-XL and GPT-J) on Reliability, Generalization, and Locality. Results are reported from \citet{yao2023editing}.}

\label{tab:ke-results}
\resizebox{\textwidth}{!}{%
\begin{tabular}{lllcccccccccccc}
\toprule
\textbf{Dataset} & \textbf{Model} & \textbf{Metric} & \textbf{FT-L} & \textbf{SERAC} & \textbf{IKE} & \textbf{CaliNet} & \textbf{T-Patcher} & \textbf{KE} & \textbf{MEND} & \textbf{KN} & \textbf{ROME} & \textbf{MEMIT} \\
\midrule

\multirow{6}{*}{\textbf{ZsRE}} 
 & \multirow{3}{*}{T5-XL} 
 & Reliability     & 20.71 & \textbf{99.80} & 67.00 & 5.17 & 30.52 & 3.00 & 78.80 & 22.51 & -     & -     \\
 &                  & Generalization & 19.68 & \textbf{99.66} & 67.11 & 4.81 & 30.53 & 5.40 & 89.80 & 22.70 & -     & -     \\
 &                  & Locality       & 89.01 & 98.13 & 63.60 & 72.47 & 77.10 & 96.43 & \textbf{98.45} & 16.43 & -     & -     \\

 & \multirow{3}{*}{GPT-J} 
 & Reliability     & 54.70 & 90.16 & \textbf{99.96} & 22.72 & 97.12 & 6.60 & 98.15 & 11.34 & 99.18 & 99.23 \\
 &                  & Generalization & 49.20 & 89.96 & \textbf{99.87} & 0.12 & 94.95 & 7.80 & 97.66 & 9.40  & 94.90 & 87.16 \\
 &                  & Locality       & 37.24 & 99.90 & 59.21 & 12.03 & 96.24 & 94.18 & 97.39 & 90.03 & 99.19 & \textbf{99.62} \\

\midrule

\multirow{6}{*}{\textbf{CounterFact}} 
 & \multirow{3}{*}{T5-XL} 
 & Reliability     & 33.57 & \textbf{99.89} & 97.77 & 7.76 & 80.26 & 1.00 & 81.40 & 47.86 & -     & -     \\
 &                  & Generalization & 23.54 & \textbf{98.71} & 82.99 & 7.57 & 21.73 & 1.40 & 93.40 & 46.78 & -     & -     \\
 &                  & Locality       & 72.72 & 99.93 & 37.76 & 27.75 & 85.09 & \textbf{96.28} & 91.58 & 57.10 & -     & -     \\

 & \multirow{3}{*}{GPT-J} 
 & Reliability     & \textbf{99.90} & 99.78 & 99.61 & 43.58 & \textbf{100.00} & 13.40 & 73.80 & 1.66  & 99.80 & 99.90 \\
 &                  & Generalization & 97.53 & \textbf{99.41} & 72.67 & 0.66 & 83.98 & 11.00 & 74.20 & 1.38  & 86.63 & 73.13 \\
 &                  & Locality       & 1.02  & \textbf{98.89} & 35.57 & 2.69 & 8.37  & 94.38 & 93.75 & 58.28 & 93.61 & 97.17 \\

\bottomrule
\end{tabular}
}
\end{table*}

\label{ssec:datasets_list}

A variety of datasets have been curated to evaluate KE methods across different tasks. These datasets assess a model’s ability to integrate new facts, correct misinformation, and retain knowledge while minimizing unintended side effects. Based on the nature of their outputs, these datasets can be categorized into \textit{generation-based} and \textit{classification-based} datasets.

\section{Detailed information about datasets}
\label{app:datasets}
\subsubsection{Generation-Based Datasets}

\paragraph{zsRE \citep{zsRE}} 
The \textbf{Zero-Shot Relation Extraction (zsRE)} dataset is widely used in KE evaluations, particularly in QA tasks. It consists of relation-specific templates sourced from Wikipedia, covering a broad range of entity-relation-object tuples. Each entry includes a valid question and an associated factual statement, with paraphrases that help test the robustness of KE methods against semantically equivalent prompts.

\paragraph{CounterFact \citep{ROME}}
\textbf{CounterFact} is designed to evaluate how well KE techniques modify a model’s underlying factual knowledge rather than merely adapting to superficial wording changes. It was introduced alongside the ROME method. Each entry is derived from ParaRel \citet{pararel} and consists of a structured knowledge triple alongside carefully crafted prompt templates. All subjects, relations, and objects originate from Wikidata, making it straightforward to assess consistency across multiple paraphrases.

\paragraph{MQuAKE \citep{MeLLo}}
\textbf{MQuAKE} is a benchmark dataset focusing on \textit{multi-hop reasoning}. It includes both counterfactual and outdated factual scenarios, requiring models to propagate edits through interconnected facts. Constructed from Wikidata, MQuAKE presents a challenging test for KE methods to verify whether updates remain consistent across related queries.

\paragraph{WikiGen \citep{MEND}} 
\textbf{WikiGen} is introduced in MEND to evaluate KE in a free-form generation setting. It consists of question-answer pairs derived from randomly sampled Wikipedia sentences, where the answers are generated using a pre-trained distilGPT-2 model. Fewer than 1\% of its samples align with the base model’s 10-token greedy predictions, making it a challenging benchmark for measuring edit reliability and generalization.

\paragraph{CKEBench}
CKEBench was introduced to address the limitations of existing KE datasets in handling commonsense knowledge expressed in natural language. Derived from ATOMIC, it covers everyday scenarios with implications like intents, reactions, and effects, framed through relations such as \texttt{xIntent} and \texttt{oEffect}. What sets CKEBench apart is its support for multiple reasoning formats—open-ended generation, multiple choice, and binary classification (true/false). This makes it a versatile benchmark for evaluating whether KE methods can edit free-text commonsense knowledge while preserving coherence and plausibility.

\paragraph{AbsATOMIC (Conceptualized Triples)}
To test whether LLMs can be edited at a higher conceptual level beyond specific instances, AbstractATOMIC was constructed by rephrasing ATOMIC’s knowledge into generalized, abstract templates using GPT-4. These conceptualized triples replace surface-level details with high-level semantic roles (e.g., “PersonX engages in enjoyable group activities”), enabling evaluations of generalization in knowledge editing. The abstraction also supports compositional reasoning and robustness to paraphrase.

\paragraph{ATOKE}
Temporal Knowledge Editing (TKE) poses a unique challenge: modifying models to reflect updated facts without erasing historically valid information. To benchmark this task, ATOKE (Assessment of Temporal Knowledge Editing) was introduced. Built from Wikidata and curated factual timelines (e.g., U.S. presidents), ATOKE tests if models can answer both present and past questions accurately across time-based edits. Each fact is timestamped, and edits evolve the model’s internal timeline, ensuring consistency across temporal transitions.

\paragraph{SafeEdit}
While detoxification has gained prominence in LLM safety research, most existing datasets target classification rather than generative reasoning. SafeEdit was designed to fill this gap. It consists of prompts in nine unsafe categories (e.g., illegal activity, self-harm) along with both safe and unsafe completions. These were generated using GPT-4 and manually curated. The dataset allows for fine-grained evaluation of whether KE methods can neutralize toxic completions without sacrificing generative fluency.

\subsubsection{Classification-Based Datasets}

\paragraph{FEVER \citep{FEVER}} 
The \textbf{Fact Extraction and Verification (FEVER)} dataset contains Wikipedia-based claims labeled as \textit{supported}, \textit{refuted}, or \textit{not enough info}. It has been adapted for KE by grouping claims on similar topics and introducing paraphrases and altered labels, providing a robust test for how well models preserve or modify factual knowledge.

\paragraph{VitaminC \citep{vitamin}} 
\textbf{VitaminC} is a large-scale fact-checking dataset derived from Wikipedia revisions, each labeled as \textit{entailed} or \textit{contradicted} by an accompanying evidence statement. It is particularly useful for testing a model’s ability to integrate factual updates without inadvertently propagating errors to unrelated claims.

\paragraph{SCOTUS \citep{GRACE}}
\textbf{SCOTUS} is adapted from a corpus of U.S. Supreme Court case documents, categorized into 11 legal topics. Due to changes in legal definitions and classifications over time, it presents a unique challenge for KE, requiring models to update domain-specific knowledge while preserving historical context.

\paragraph{ConceptEdit}
ConceptEdit focuses on a novel form of KE: modifying conceptual definitions (e.g., animal taxonomy) and observing their impact on instance classification. The dataset was built using DBpedia and Wikidata by selecting concepts (like “Camelidae”) and associating them with natural language definitions and instance lists. When a concept’s definition is edited, models must infer which instances still belong. ConceptEdit thus evaluates the downstream semantic consequences of edits.

\paragraph{PROBE SET (MEMITCSK)}
To explore whether KE generalizes across surface form and reasoning depth, PROBE SET was created. Based on commonsense datasets like PEP3k and 20Q, it includes true/false statements with paraphrased, contradictory, and entailment-related variations. This setup tests whether knowledge edits propagate semantically across related linguistic structures. The evaluation is grounded in binary judgments (true vs. false), positioning PROBE SET as a classification-based resource.

\paragraph{Wikibias}
Addressing the growing concern of social biases in LLMs, Wikibias offers a benchmark for stereotype editing. Extracted from real Wikipedia content, the dataset pairs biased and unbiased factual claims involving professions, gender, race, and other social roles. Each example allows comparison of the model’s preference toward stereotypical vs. neutral formulations. By design, Wikibias targets binary classification of bias presence and factual validity.

\paragraph{BiaScope}
BiaScope was constructed to evaluate KE methods on fine-grained social bias mitigation. It merges data from StereoSet and CrowS-Pairs with GPT-4-generated paraphrases and human annotations. The dataset contains two parts: (1) non-biased commonsense knowledge that must be preserved, and (2) stereotype-laden sentences that should be edited. This dual-purpose setup enables controlled testing of bias removal without degrading general knowledge.

\paragraph{RelEdit}
RelEdit was constructed to evaluate conceptual knowledge editing, with a specific focus on the model's relational reasoning capabilities after an edit. The benchmark is built upon the DBpedia ontology and contains a hierarchy of concepts and their corresponding instances. The dataset is structured to assess the "ripple effects" of an edit at two levels: (1) the instance level, evaluating changes in the relationships between a concept and its instances, and (2) the concept level, evaluating changes among related concepts. This two-level setup enables a comprehensive assessment of whether an edit has been deeply integrated into the model's knowledge structure, going beyond simple definition recall.

\end{document}